# Squeeze-SegNet: A new fast Deep Convolutional Neural Network for Semantic Segmentation

[1]Geraldin NANFACK, [2]Azeddine ELHASSOUNY, [3]Rachid OULAD HAJ THAMI
IRDA Team, ADMIR LAB, RABAT IT CENTER,
ENSIAS, University of Mohamed 5 in Rabat, Morocco
[1]nanfackg@gmail.com,  [2]elhassounyphd@gmail.com, [3]rachid.ouladhajthami@gmail.com

**ABSTRACT**

The recent researches in Deep Convolutional Neural Network have focused their attention on improving accuracy that provide significant advances. However, if they were limited to classification tasks, nowadays with contributions from Scientific Communities who are embarking in this field, they have become very useful in higher level tasks such as object detection and pixel-wise semantic segmentation. Thus, brilliant ideas in the field of semantic segmentation with deep learning have completed the state of the art of accuracy, however this architectures become very difficult to apply in embedded systems as is the case for autonomous driving. We present a new Deep fully Convolutional Neural Network for pixel-wise semantic segmentation which we call Squeeze-SegNet. The architecture is based on Encoder-Decoder style. We use a SqueezeNet-like encoder and a decoder formed by our proposed squeeze-decoder module and upsample layer using downsample indices like in SegNet and we add a deconvolution layer to provide final multi-channel feature map. On datasets like Camvid or City-states, our net gets SegNet-level accuracy with less than 10 times fewer parameters than SegNet.

## 1. INTRODUCTION

The scene understanding [1] from images is and remains a worrying subject in computer vision. Since the era of pioneers in this field with LeNet [2] and its first application in convolutional neural network in Computer Vision, researchers threw themselves in vain before AlexNet [3] Architecture appeared improving significatively advances in Deep Learning with classification task of images. Indeed, the degree of learning and understanding scene by computer will be higher if one succeeds in the detection of objects [4]. J. GrawaAl et al. [5] go far to say that semantic segmentation takes the state of object detection and deals with detecting the exact region that object occupies. The problem of semantic segmentation occurs when we want to pass to a  higher level [6] by making dense prediction on pixel. Thus, image semantic segmentation is a classification task consisting to design a method in other to assign a meaningful label to each pixel of the image in which each label represents the class of the detected object.  This task, not so easy even for a human, has a major complexity for a computer due mainly to problems like the variation of pose and appearance of the scene [7]. Because of the fact that it has its strong application in robotics, satellite imagery for relief mapping [8], autonomous driving [9], automatic annotation of images especially in the field of medical imaging [10], it is a subject that make a lot of attention the recent past years. Therefore to solve problems of low level of the image in computer vision, methods developed have been successful [11] thanks to application of the popular machine learning algorithms, popularity especially unheard of by the advent of Deep Learning [12]. Although Deep Convolutional Neural Networks (DCNNs) have pushed the performance of computer vision systems to soaring heights on a broad array of high-level  problem including image classification [13] and object detection [14] with AlexNet [15] , ZFNet [16], VGGNet [17], GoogLeNet [18], ResNet [19], MobileNet [20],  where DCNNs trained end-to-end manner have delivered strikingly better results than systems relying on hand-crafted features, it has been well used in low-level tasks such as the pixel wise prediction of semantic segmentation, benefiting from the power of DCNNs in capturing abstract features and invariances [21]. Thus, several DCNNs for semantic segmentation have been developed using the power of previous architectures initially designed for the classification task. Aligning themselves to this principle, several methods were then devised. However, for the most part, they follow DCNN lackness and failures such as the size of the network and the time consumed for computational cost, which disadvantages their use for embedded applications, in particular for semantic segmentation for autonomous driving. We then address this paper in the goal to deal with these problems which are size of the network and the time consumed for computational cost. We develop a new fully convolutional architecture for the image semantic segmentation especially images of cities, achieving accuracy similar than the successful SegNet [11] architecture and has 4 times fewer parameters. In the next sections, after presenting a state of the art of existing architectures, we will present our new technique implemented in caffe framework [22] and by the end we will make a benchmark of results.

## 2. RELATED WORK

As mentioned previously, semantic segmentation has been a preoccupying topic in recent years. The methods are divided into two main groups: hand-crafted features and deep learning. Methods that relied on hand-crafted features were use machine learning algorithms to perform classification task such as Boosting [23], [24], Random Forests [25], or Support Vector Machines [26]. Some improvements have been achieved by incorporating richer information form context [27] and structured prediction techniques [28], [29], [30], [31], but the performance of these systems has always been compromised by the need for improved and powerful features for classification. Due to the limitations and weakness of hand-crafted methods, and thanks to the scientific community and its large datasets like Imagenet [32] without sparing the great machine computational capacities, convolutional neural networks (CNNs) which in the past were had little applied could give a new breath [33] in computer vision. In fact his great capability to learn abstract and powerful features [21] comes with the very first resounding success in classification of images and the object detection. This success of DCNNs for object classification has recently inspired researchers to exploit these learned features for structured prediction problems like segmentation [11]. One of the most successful state-of-art was the Fully Convolutional Network (FCN) by Long et al [34]. The insight of that approach was to take advantage of existing CNNs as powerful visual models that are able to learn hierarchies of features. The transformed those existing and well-known classification models AlexNet [3], VGG (16-layer net) [17] and GoogLeNet [18] into fully convolutional ones by replacing the fully connected layers with convolutional ones to output spatial maps instead of classification scores. Then those maps are upsampled using fractionally strided convolutions (deconvolutions [16]) and bilinear interpolations [35] to upsample and produce dense predictions for semantic segmentation with input of arbitrary sizes. The FCN is considered a milestone [36] since it showed how CNNs can be trained end-to-end to make dense predictions for semantic segmentation with inputs in arbitrary sizes. However, its efficiency is still far from real-time execution at high resolutions because at the network grows no further increase performance is observed and it still a high size model (>200M). To overcome the problem of stride convolution, Hyeonwoo Noh et al. a segmentation algorithm (DeconvNet) [37] by learning a deep deconvolution. Apart form the FCN architecture and DeconvNet, other variants were developed to transform a ntwork whose purpose was classification (Encoder) to make it suitable for segmentation by adding a Decoder part [37] before removing fully connected layers. SegNet [11] is a clear example of this divergence. It is an Encoder-Decoder style and has an encoder network identical to the 13 convolutional layers in VGG16 network [17]; also a decoder network is still a serie of convolutional layers but his novelty lies in the manner in which the decoder upsamples its corresponding encoder to perform non-linear upsampling by using the encoder downsample indices [11]. However due to the lackness of convolutional architecture to the capture of global context of object and inherent invariance to spatial transformations, an approach was proposed to apply a post-processing stage using a Conditional Random Field (CRF). So DeepLab models [38], [39] use the fully connected pairwise CRF by Krähenbühl and Koltun [40] as a separated post-processing step in their pipeline to refine the segmentation result. Another significant work applying a CRF to refine the segmentation of a FCN is the CRFasRNN by Zheng et Thal. [41]. Their main contribution of that work was the reformulation of the dense CRF with pairwise potentials as an integral part of the network. They integrate CRF with a FCN and train the whole end-to-end using the reformulation of CRFs as Recurrent Neural Networks (RNNs) [36]. To overcome the problem of elements with zeros during upsample in some architectures, some people used dilated convolutions and the most important works that used it is the multi-scale context aggregation module by Yu et al. [42], also the real time network ENet [43]. Another remarkable work for adding context information to FCN is feature fusion. This technique consists of merging a global feature with a more local feature map extracted from a subsequent layer [36]. This approach is taken by ParseNet [44] in their context module and continued by Pinheiro et al. with SharpMask [45] network mainly focused on instance segmentation. Another way to capture global context is the use of power of RNNs which are able to successfully model global contexts and improve semantic segmentation [36]. Based on the model ReNet [46] for image classification, Visin et al. proposed ResSeg [47] for a semantic segmentation task. Particular approaches have also emerged on particular datasets such as RGB-D data with Ma et al. [48] proposing an RGB-D SLAM technique based on FuseNet [49] and 3D data with the pioneer PointNet [50] composed by a subnetwork that takes a point cloud and applies a set of transforms and Multi Layer Perceptrons (MLPs) to generate features [36].

## 3. SQUEEZE-SEGNET

### 3.1 The Architecture

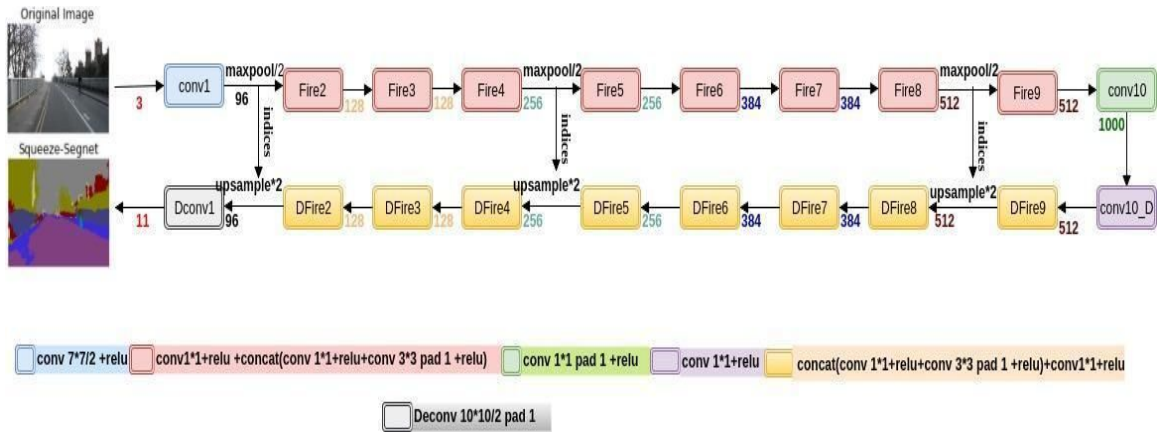

Figure 1. Architecture of Squeeze-SegNet.

As we see in the Fig. 1 Squeeze-SegNet is an encoder-decoder style. The encoder consists essentially on the SqueezeNet architecture [51]. We have chosen it to the detriment VG16-Net because these authors that are Forrest N. Iandola et al. succeeded in modeling an image classification architecture whose model size is less than 5MB. For the decoder part, we designed our own symmetrical way to the decoder, and we inversed downsample by upsample layers developed by SegNet [11].

Let's get to the encoder. It is composed by a first convolutional layer of filter kernels of size 3 with stride 2 and without padding. It was designed in SqueezeNet [51] like that in order to get a large receptive field in the first layer of network. The Fire module: it is the novelty of SqueezeNet image classification architecture; it is composed by a *squeeze* convolution layer (1x1 filters) use to reduce parameters, feeding into a concatenation of 2 *expand* layers: one of 1x1 convolution filters and another 3x3 convolution filters used to decrease the number of input channels to 3x3 filters [51]. So we used theses fire module and remove the average pooling layer of SqueezeNet which was the activation score for object classes. We were inspired by SegNet [11] which removes all the fully connected layer of VGG16-Net [18]. The decoder part: this is our proposed method to inverse the fire and convolutional layers of SqueezeNet architecture. Firstly for the last convolutional layer (conv10 on Fig. 1), form [16], we can use another convolutional layer to inverse conv10; it is what we have done. Also to inverse Fire module, we design a DFire module as a serie of concatenation of *expand* module of SqueezeNet followed by a *squeeze* module. And the end, to invert the first convolutional layer (conv1) we were inspired by the work of H Noh [16] on DeconvNet who noticed that the deconvolution layer can play many roles in computer vision and the more important are their ability to upsample and produce enlarge and dense activation map. It was exactly what we needed because the first conv1 decoder layer has a stride of 2; it means that it reduces the spatial dimension of input image. By the end, it is important to note that we inverse downsample by upsample layer of SegNet [11] with the principle to share maxpool indices from downsample and upsample layers.

## 3.2 Details of Implementation

As we said in the above section, we designed the Squeeze-SegNet architecture for semantic segmentation, a preview is in the Fig. 4. We train this model on the caffe deep learning framework [52]. It took approximately 9 hours on the CamVid [54] dataset, on the material of GPU NVidia GTX 1050. From the 35, 000th iteration, the model seemed to converge and the learning was slower and slower due to the major problem of backpropagation [53] on neural network: vanishing gradient problem [54]. So we ended the learning after 44,000 iterations (Fig. 3), we can see it in the figure . The optimisation method of the Gradient descent with mini batch of 4 With the table of Fig. 2, we show that our net has around 2.8 billions of parameters and the Fig. 3 proves squeeze-segnet learn very fastly.

| Layer name | Output size | filter size | Number of parameters |
|---|---|---|---|
| input | 480x360x3 | | |
| conv1 | 237x177x96 | 7x7/2 (x96) | 14208 |
| maxpool1 | 118x88x96 | 3x3/2 | 0 |
| Fire2,Fire3, | 118x88x128 | | 11920+12432 |
| Fire4 | 118x88x256 | | 45344 |
| maxpool4 | 59x44x256 | 3x3/2 | 0 |
| Fire5 | 59x44x256 | | 49440 |
| Fire6,Fire7 | 59x44x384 | | 104880+111024 |
| Fire8 | 59x44x512 | | 188992 |
| maxpool8 | 29x22x512 | 3x3/2 | 0 |
| Fire9 | 29x22x512 | | 197184 |
| conv10 | 31x24x1000 | 1x1/1 (x1000) | 513000 |
| conv10_D | 29x22x512 | 1x1/1(x512) | 512512 |
| DFire9 | 29x22x512 | | 197184 |
| upsample8 | 59x44x512 | | 0 |
| DFire8, DFire7 | 59x44x384 | | 188864+111024 |
| DFire6, DFire5 | 59x44x256 | | 104752+74032 |
| upsample4 | 118x88x256 | | 0 |
| DFire4, DFire3 | 118x88x128 | | 45216+12432 |
| DFire2 | 118x88x96 | | 12432 |
| upsample1 | 237x177x96 | | 3760 |
| conv1_D | 480x360x11 | 10x10/2(x11) | 203637 |
| Total of parameters | | | 2,714,269 |
| Size of model (x32bits if its the codage) | | | 10.35MB |

Fig2: Number of parameters of Squeeze-SegNet

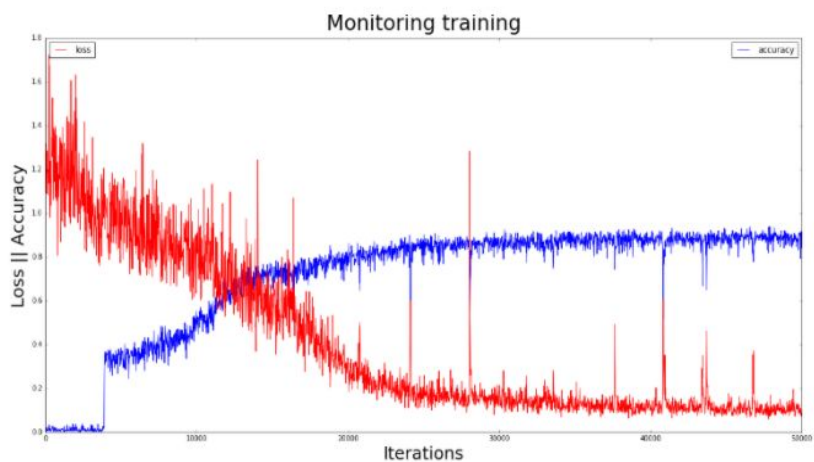

Fig3: Monitoring training of Squeeze-SegNet

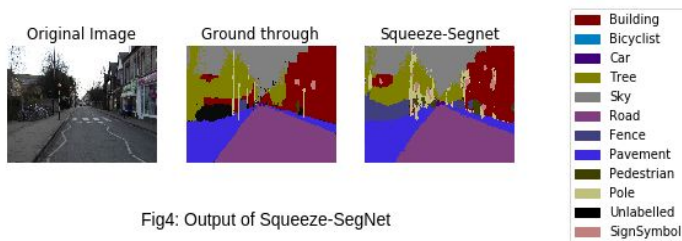

Fig4: Output of Squeeze-SegNet

## 3.3 Results

As we said above, we trained our net with CamVid [53] dataset of the version of 11 classes (table 1). In this figure, it an output of Squeeze-SegNet on a testing data. We can see the matching between colors and objects detected and segmented scene of autonomous driving. We also provide in the table 2 a first release of accuracy per classes because we will get more interesting results based on implementations details of dataset. But this a vanguard of first release of squeeze-segnet in terms of results.

Table 1. Accuracy of Squeeze-SegNet on CamVid data

| Objects | sky | Building | Pole | Road | Side-Walk | Tree | Sign | Cars | Fence | Pedestrian | Bicyclist | Average Accuracy |
|---|---|---|---|---|---|---|---|---|---|---|---|---|
| Accuracy | 0.945 | 0.889 | 0.369 | 0.936 | 0.936 | 0.755 | 0.198 | 0.01 | 0.977 | 0.644 | 0.676 | 0.667 |

In view of these results, it seems clear that the net does not manage to learn well how to segment 2 objects; we can think that it is due to the form of object and the problem of imbalanced classes because each object has not the same percentage of a pixels belonging to it. To overcome with this problem, we implement another variant in which during training we look particularly of this problem.

### 3.4 Benchmarking

We benchmark Squeeze-SegNet with other architectures and their accuracy and especially their type of neural network and their model size. In table 3 we compare the performance of Squeeze-SegNet with existing state-of-the-art algorithms on CamVid dataset. It is important to notice the fact that for semantic segmentation, the type of neural network is very important because for example only convolutional has the great advantage to have in the input of the net image with any spatial shape, others types have not got this advantage.

Table 2. Results on CamVid test set of SegNet-Basic, SegNet, ENet and Squeeze-SegNet

| Architecture | Type | Parameters | Model size(fp32) | Class Average Accuracy |
|---|---|---|---|---|
| SegNet | Convolutional | 29.4 M | 117.8 MB | 0.65 |
| SegNet-Basic | Convolutional | - | - | 0.63 |
| ENet | Residual | 0.36 M | 1.5 MB | 0.68 |
| **Squeeze-SegNet** | **Convolutional** | **2,7M** | **10.35MB** | **0.667** |

## 4. CONCLUSION

At the end of our work, we proposed Squeeze-SegNet, which is a model having 4 and more times less than the size of the architectures and the computational time of DCNN for the semantic segmentation of the images. parameters than SegNet and having similar efficiencies. It is an architecture of Fully convolutional network that can have inputs of any size and intended for the best for the segmentation of scenes of images of cities or can even generalize elsewhere. Despite the difficulties of our study based essentially on the technical conditions, we propose nevertheless by this new Deep convolutional architecture, a simple and efficient method to segment the images of city. Trained on the framework of deep learning caffe, it is a model whose pretrained weighs around or less than 20MB without any operation. In spite of its power, other works in this framework will be able to register to improve it.

## ACKNOWLEDGES

We are thankful to the program of European Union ERMIT (Entreprenariat, Ressources, Management, Innovation et Technologies) for the scholarship awarded to the first author without which this work would not have taken place. Thanking the Lord God, We would also like to thank the coordination of the Research Master in Data Sciences and Big Data of ENSIAS and then in another measure the family and the future Doctor Harry Kamdem for support.